\documentclass[10pt,twocolumn,letterpaper]{article}

\usepackage{iccv}
\usepackage{times}
\usepackage{epsfig}
\usepackage{graphicx}
\usepackage{amsmath}
\usepackage{amssymb}

\usepackage{graphicx}
\usepackage{amsmath}
\usepackage{amssymb}
\usepackage{booktabs}

\usepackage{amsmath, bm}
\usepackage{bbm}
\usepackage{multicol}
\usepackage{multirow}

\usepackage[pagebackref=true,breaklinks=true,letterpaper=true,colorlinks,bookmarks=false]{hyperref}

\usepackage[capitalize]{cleveref}
\crefname{section}{Sec.}{Secs.}
\Crefname{section}{Section}{Sections}
\Crefname{table}{Table}{Tables}
\crefname{table}{Tab.}{Tabs.}

\def\eg{\textit{e.g.}}
\def\etal{\textit{et al.}}

\def\ie{\textit{i.e.}}

\iccvfinalcopy 

\def\iccvPaperID{5612} 

\ificcvfinal\fi

\begin{document}

\title{PourIt!: Weakly-supervised Liquid Perception from a Single Image \\ for Visual Closed-Loop Robotic Pouring}

\author{Haitao Lin\quad Yanwei Fu\quad Xiangyang Xue\\
Fudan University
}


\maketitle

  
 

\ificcvfinal\thispagestyle{empty}\fi

%
\begin{abstract}
Liquid perception is critical for robotic pouring tasks. It usually requires the robust visual detection of flowing liquid. However,
while recent works have shown promising results in liquid perception, they typically require labeled data for model training, a process that is both time-consuming and reliant on human labor.
To this end, this paper proposes a simple yet effective framework PourIt!, to serve as a tool for robotic pouring tasks. We design a simple data collection pipeline that only needs image-level labels to reduce the reliance on tedious pixel-wise annotations. 
Then, a binary classification model is trained to generate Class Activation Map (CAM) that focuses on the visual difference between these two kinds of collected data, \ie, the existence of liquid drop or not. 
We also devise a feature contrast strategy to improve the quality of the CAM, thus entirely and tightly covering the actual liquid regions. Then, the container pose is further utilized to facilitate the 3D point cloud recovery of the detected liquid region. Finally, the liquid-to-container distance is calculated for visual closed-loop control of the physical robot. 
To validate the effectiveness of our proposed method, we also contribute a novel dataset for our task and name it PourIt! dataset. Extensive results on this dataset and physical Franka robot have shown the utility and effectiveness of our method in the robotic pouring tasks. Our dataset, code and pre-trained models will be available on the project page~\footnote{Project page. \url{https://hetolin.github.io/PourIt}}.
\end{abstract}

 \begin{figure}[htb]
    \centering
    \includegraphics[width=1\linewidth]{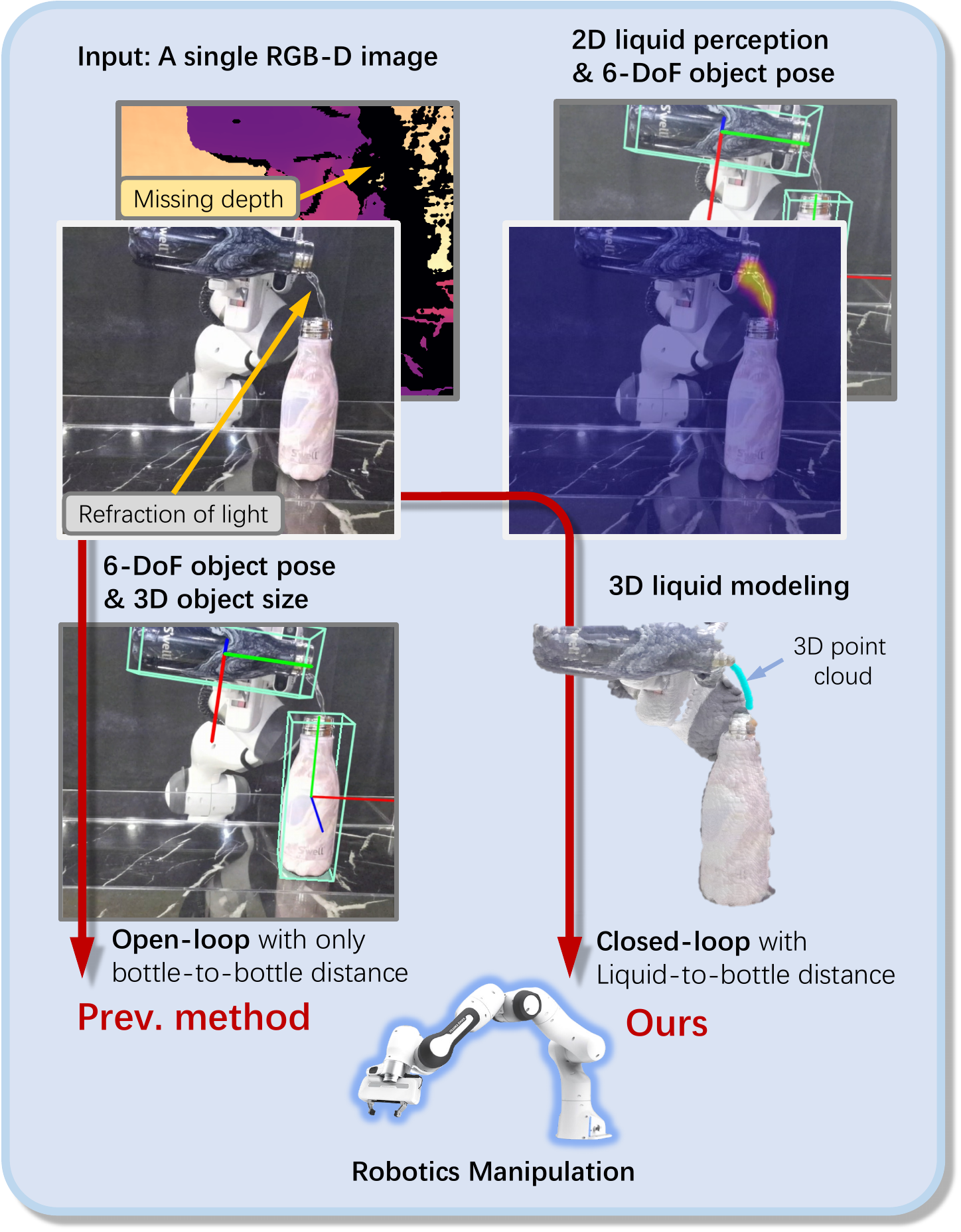}
    \caption{Our visual closed-loop robotic pouring. Unlike ~\cite{lin2022sar}, our approach recovers a 3D point cloud of the liquid using the source container's pose and 2D liquid perception data. This allows the robot to pour accurately based on visual feedback, even without the depth measurement of the liquid.
    }
        \vspace{-0.1in}
    \label{fig:teaser}
    \vspace{-0.1in}
\end{figure}


\section{Introduction}
Recent advances have seen the great progress in the capabilities of grasping and manipulating the rigid objects. However, the manipulation of non-rigid objects such as liquids, cloth, and rope remains a formidable challenge due to the absence of fixed patterns and geometrical shapes associated with flexible properties.
Perceiving images of liquids is particularly challenging due to their reliance on refraction of light as the primary visual cue and the absence of depth measurement as shown in Fig.~\ref{fig:teaser}~(top left).
%
%
%
Thus, further research aimed at enhancing the ability to perceive liquids would be highly beneficial, particularly for robots engaged in real-world service tasks such as cooking, drink serving, and plant watering.



Pouring water is a highly relevant task in the field of liquid manipulation, yet it poses a number of significant challenges. These challenges include (1) the need for large amounts of pixel-wise annotated data to facilitate effective training, (2) the lack of salient visual cues within images, and (3) the unavailability of a reliable depth measurement system for accurate liquid pouring. Addressing these challenges is crucial for advancing the state of liquid manipulation technology and enabling practical applications in domains such as robotics and industrial automation.

To address the aforementioned challenges, researchers have explored the use of extra sensors to generate ground-truth labels for robotics pouring. For example, Schenck \etal~\cite{schenck2017visual} skillfully utilize the thermal camera and heating water to obtain the ground-truth annotations.  However, this approach is time-consuming and relies on additional equipment. Other methods~\cite{liang2019making, wilson2019analyzing, xompero2022audio, ishikawa2021audio, liu2021va2mass, wang2022improving} rely on the audio but not visual signals to help robotics pouring, which limits their efficacy in noisy environments. Lin~\etal~\cite{lin2022sar} simply utilize the estimated container's pose and size to calculate the initial pouring point without any liquid perception (Fig.~\ref{fig:teaser}), thus cannot guarantee whether the liquid is poured into target container in a closed-loop manner. More recently, a self-supervised method~\cite{narasimhan2022self} has been proposed to transfer colored liquid into transparent liquid without the need for manual annotations.
However, this approach is limited by its reliance on colored liquid and a statically placed transparent container, which limits its applicability in more general environments.
Therefore, the challenge of robustly perceiving dynamically out-flowing liquid during pouring operations remains an open question that requires further research.

 

This paper presents a simple yet effective framework that solves challenges in robotic pouring tasks, particularly during the pouring stage. Our open-source library can complement existing tools such as  MoveIt!~\cite{chitta2012moveit} for motion planning and GraspIt!~\cite{miller2004graspit}  for grasping planning. We address three specific challenges, including \textit{limited annotated real-world data for training, non-salient visual cues of liquids, and unavailable depth measurement of the liquid.} 
(1) To address the first challenge, we design an semi-automatic pipeline that will collect many images with simple two-class labels indicating the existence of flowing liquid from the bottleneck of source container. This pipeline enables the robot to collect more data, thereby improving the performance of the network.
(2) To tackle the second challenge,
we classify the images by distinguishing between the two types of collected data (existence of out-flowing liquid or not). However, such a classification task may not cover the liquid entirely. To overcome this, we use a feature contrast strategy to pull foreground local features close and separate foreground-background local features to identify the liquid region.
%
%
(3) For the third challenge, we make the gravitational assumption that the stream of liquid always aligns along the body of the container and thus approximate the 3D shape of the liquid by using estimated container pose. We recover 3D point cloud of the liquid as visual feedback for robot control.

\begin{figure}[b]
    \centering
    \includegraphics[width=0.95\linewidth]{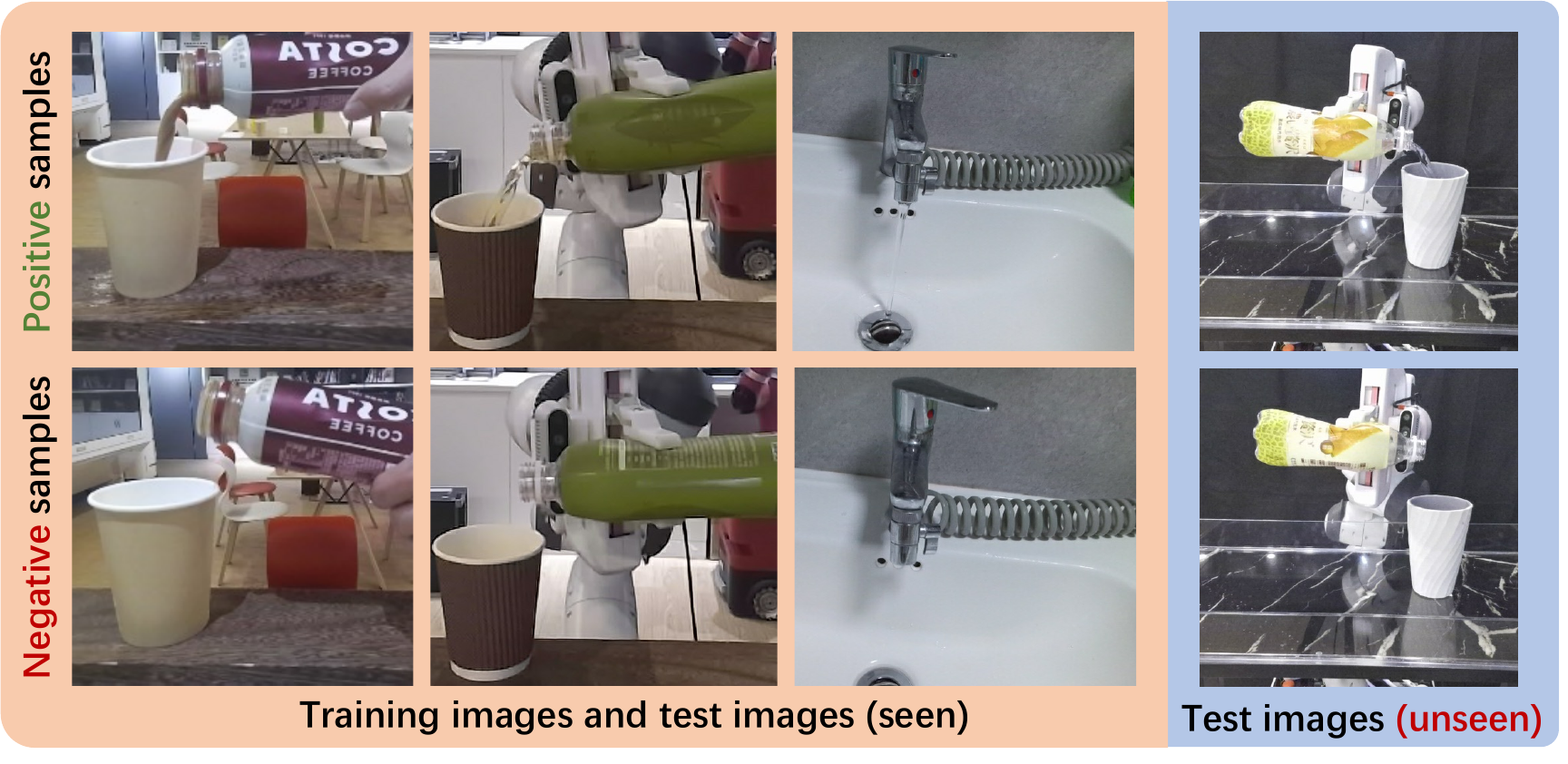}
    \caption{The exemplar of our proposed PourIt! dataset. It includes images with liquid and non-liquid, totaling 3354 training images, 374 seen test images, and 336 unseen ones.  \label{fig:dataset_example}}
   
\end{figure}

Technically, we first collect two types of real-world data by robot or human executing pouring action with using empty or full container, and turning on or off the taps as shown in Fig.~\ref{fig:dataset_example}. The data is divided into positive samples if there is flowing liquid, otherwise negative samples.
Then we train a classification network to distinguish the positive and negative samples, which drives the network to focus on the difference of the two types of data, \ie, existence of liquid or not. Thus the derived CAM will coarsely focus on the liquid regions. We further utilize this initial CAM to separate the feature maps into foreground and background local features. Then, we bridge the distance between foreground-foreground pairs while widening the distance between foreground-background pairs. This generates a better CAM which aligns with low-level boundaries and completely covers the target liquid region. Then, we use the off-the-shelf category-level pose estimation, \eg, SAR-Net~\cite{lin2022sar} to recover the pose of the source container for calculating the plane equation aligned with gravity and orientation of container's neck. Finally, the 3D point cloud of liquid is approximated by calculating the line-plane intersections, where the lines are the rays back-projected from estimated 2D liquid region.

Im summary, the main contributions of this paper are: (1) We first propose a novel weakly-supervised pipeline to transfer the 2D liquid perception problem into a classification task. We also propose a feature contrast strategy to improve quality of CAM. To the best of our knowledge, we are the first work to perceive and model the 3D liquid from a single image without temporal information.  (2) We first propose the method of approximate 3D shape of liquid  by utilizing the 6-DoF pose of source containers and estimated liquid mask for robotic pouring-related task. (3) We deploy our real-time framework (10Hz) on the physical Franka robot to serve as a visual feedback to pour liquid more accurately into target container. (4) We also propose the PourIt! dataset, which could be a test bed benchmark of the self/weakly-supervised liquid segmentation task for computer vision and robotics community.
We believe that PourIt! framework should endow the ability for robots in continuous self-supervised learning. For example, extending the tasks in Fig.~\ref{fig:teaser} to continuously make the robot mutually pour liquid from two containers to collect data in a self-supervised manner, thus fine-tuning the model using more data for better liquid perception.

\section{Related Work}
\noindent\textbf{Liquid Perception.}
Accurately perceiving the liquid is challenging as the lack of fixed patterns of geometry shape.  The simulator~\cite{schenck2017reasoning} that can synthesize the fluid is a good choice to generate an amount of labeled data to train a model. But the real-synthetic image domain gap usually degrades the performance of the synthetically-trained models. Thus our method directly uses the real-world for training which avoids the so-called reality gap.
Recent approaches~\cite{schenck2017visual, schenck2018perceiving, schenck2016towards}  skillfully use the thermal camera aligned with the color image to generate the accurate ground-truth label for transparent liquid perception. However, the extra calibrated thermal sensor and heating water requirement are usually cost-consuming and human-labor intensive. 
Another line of works utilize the specialty of liquid motion, such as the optical flow~\cite{yamaguchi2016stereo} or the sound~\cite{liang2019making, wilson2019analyzing, xompero2022audio, ishikawa2021audio, liu2021va2mass, wang2022improving} to perceive the liquid.
For example, Yamaguchi~\etal~\cite{yamaguchi2016stereo} adopt the optical flow of the liquid motion to perceive the water. However, this method demands no redundant motion when executing the pouring action, as the optical flow is sensitive to the background change, which will easily introduce the outlier pixels. Another works~\cite{liang2019making, wilson2019analyzing} capture and analyze the cue of the audio vibration to estimate the weight or height of liquid in the container. These methods cannot be applied to perceiving the regions of liquid drop. 
Narasimhan~\etal~\cite{narasimhan2022self} first propose the self-supervised method by taking generative model CUT~\cite{park2020contrastive} to transform the color liquid into transparent one, further facilitating background subtraction for training the segmentation model. But it is only suitable for liquid detection inside static transparent containers.
Our proposed method employs simple image-level labels to weakly supervise the model for perceiving and estimating the 3D shape of the regions of liquid drop. Thus, our proposed method can reduce the cost of either human-labor or equipment.


\noindent\textbf{Weakly-supervised Semantic Segmentation.} Weakly-supervised Semantic Segmentation (WSSS) usually uses coarse labels~\cite{ahn2018learning, kolesnikov2016seed, khoreva2017simple, papandreou2015weakly, bearman2016s, dai2015boxsup, lin2016scribblesup} for supervision but produces pixel-level localization maps. The image-level supervised WSSS approaches usually take the CAM technique~\cite{zhou2016learning, selvaraju2017grad} to train a classification network, and thus generate the initial pseudo pixel-level labels for supervision. The CAM only focuses on the most discriminative regions but not all parts of the objects. Some methods address this problem by using saliency maps~\cite{lee2021railroad, sun2019saliency}, erasing strategy~\cite{sun2021ecs, wei2017object, zhang2021complementary, chen2022class}, or accumulation strategy~\cite{jiang2019integral, kim2021discriminative, yao2021non} to help the CAM attend to more complete objects. Recent Transformer architecture has proven superior performance in semantic segmentation tasks~\cite{cheng2021per, xie2021segformer, zheng2021rethinking}. In the field of WSSS, for example, AFA~\cite{ru2022learning} exploits the semantic affinity from multi-head attention in Transformers and pixel-adaptive refinement that incorporates low-level image appearance information to refine the pseudo labels.
However, these methods cannot accurately localize the liquid, especially for the transparent liquid, due to non-salient visual cues of boundaries.

\section{Methodology}
 \begin{figure*}[h]
    \centering
    \includegraphics[width=1\linewidth]{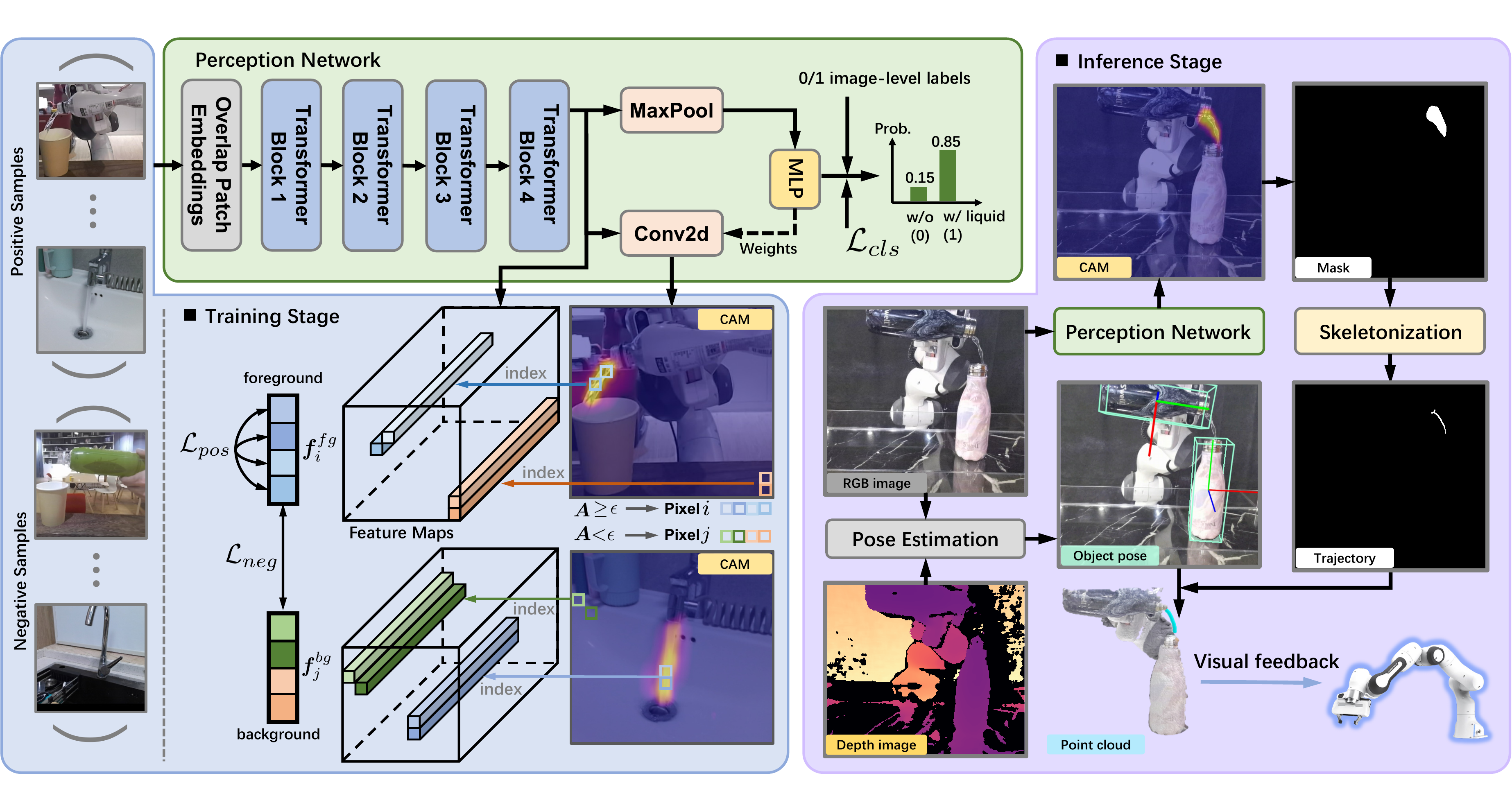}
    \vspace{-0.1in}
    \caption{The workflow of our proposed PourIt! framework. Given the positive and negative samples of images with simple 0/1 image-level labels, we first use the Transformer backbone to extract features and the MLP layers to predict the image classes. Then, the derived CAM is used to index the localization of foreground and background pixels of the feature maps by using the threshold $\epsilon$. In order to improve the quality of CAM,  we force the network to pull close the foreground features while pulling apart the background ones. At the inference stage, we use a pose estimation network to recover the 6-DoF poses of source and target containers and a perception network to estimate the potential liquid mask from derived CAM. The trajectory of the liquid is then extracted according to the morphological shape of the mask. Finally, the point cloud of liquid is reconstructed by using the predicted 6-DoF object poses and 2D liquid trajectory, which provides real-time visual feedback of liquid-to-container distance as high-level guidance of robot control.\label{fig:pipeline}}
    \vspace{-0.15in}
\end{figure*}

\noindent\textbf{Task overview.} This paper focuses on perceiving  dynamically flowing liquid when pouring into the target container. This task is very challenging, as the target container maybe moving or its neck is narrow.
Such a task requires real-time visual and control feedback to accurately control the liquid motion to avoid liquid spilling out of the target container. 

\noindent\textbf{Problem Formulation.} Given a monocular RGB image $\mathcal{I} \in \mathbb{R}^{H\times W \times 3}$ of the pouring liquid, our goal is to estimate the pixel-wise segmentation $\mathcal{M} \in \mathbb{R}^{H\times W \times 1}$ of a stream of liquid which is poured out from the source container's neck. This paper supposes that the 6-DoF rigid pose transformation $\{\bm{R};\bm{t}\} \in SE(3)$  of the source container is known beforehand with the 3-DoF rotation $\bm{R} \in  SO(3)$ and 3-DoF translation $\bm{t} \in \mathbb{R}^3$. 
Due to gravity, the direction of liquid is always kept consistent with the plane that aligned with the orientation of the container’s neck and gravity. Thus, given the camera intrinsic, we finally recover the approximate 3D point cloud $\bm{P}$ by using the pose of container $\{\bm{R};\bm{t}\}$ and estimated liquid mask $\mathcal{M}$.

\subsection{Data Collection}
As there are no available mask annotations of liquid in our experiment, thus it is essential to guide the network to focus on the region of liquid drop. Inspired by the idea of weakly-supervised segmentation, we aim to utilize the class activation map derived from the network for liquid segmentation. 
Lin et al.~\cite{lin2022sar} tackled the task of pouring solid objects (beans) from the source into the target container. Using their method, we can collect much data of pouring actions.
%
%
Then we design the semi-automatic pipeline to collect two kinds of samples. 

Particularly, (1) we will first construct the $N$ positive samples $<I_1^{+}, \cdots, I_N^{+}>$ with the stream of liquid by the robot or human executing pouring action by using liquid-fulled container as in Fig.~\ref{fig:dataset_example} (column 1-2). During the stage of the source container horizontal with the ground, we record and save the RGB frames where the liquid drop always exists. (2) Then, we replicate the same action but use the empty source container, to collect the negative samples $<I_1^{-}, \cdots, I_N^{-}>$. 
 We also add other images by turning on or off the tap for data augmentation Fig.~\ref{fig:dataset_example} (column 3). Finally, 
To balance the numbers of positive and negative samples, we finally sample 1677 images for each part individually. 

\subsection{Transformer Encoder}
Our liquid perception framework uses Transformer as the backbone and an MLP layer to classify the images as in Fig.~\ref{fig:pipeline}. Given an RGB image $\mathcal{I} \in \mathbb{R}^{H \times W \times 3}$, we first split it into multiple $h \times w$ patches of size $4 \times 4$. Then these patches are further flattened and linearly projected into $h \times w$ tokens. The Transformer layer comprises a multi-head self-attention block that outputs $n$ attention maps. For each self-attention head, the patch tokens are projected into queries, $Q_i \in \mathbb{R}^{hw \times d_k}$, keys $K_i  \in \mathbb{R}^{hw \times d_k}$, and values $V_i  \in \mathbb{R}^{hw \times d_v}$, $i=1,2, \cdots, n$.
Then the attention map is estimated by the scaled dot-product mechanism as follows,
\begin{equation}
    Attention(Q_i,K_i,V_i)= Softmax(\frac{Q_i K_i^\top}{\sqrt{d_k}})V_i
\end{equation}
where $\sqrt{d_k}$ is the scalar factor; $d_k$ and $d_v$ are the feature dimension of the keys and values, respectively. 
Then we concatenate all attention maps and feed them into the feed-forward layers to obtain the feature maps. Similar to ~\cite{ru2022learning}, we use four stacked Transformer blocks in our case, thus outputting the multi-level feature maps $\bm{F}^{(i)}, i=1,2,3,4$ at $\{\frac{1}{4}, \frac{1}{8}, \frac{1}{16}, \frac{1}{16}\}$ of the original image resolution.

\subsection{CAM Generation}
After obtaining the feature maps, we apply a pooling operation on features maps $\bm{F}^{(4)} \in \mathbb{R}^{h'w'\times d}$ that outputs from the last transformer block, and then send it into Multi-Layer Perception (MLP) layers for binary classification. We use the binary cross-entry loss $\mathbb{BCE}(\cdot , \cdot)$ as the supervision,
\begin{equation}
    \mathcal{L}_{cls} = \mathbb{BCE}(\hat{y}, y)
\end{equation}
where $\hat{y}$ and $y$ are the predicted and ground-truth image-level label. At the stage of inference, we generate the Class Activation Map (CAM) $\bm{A}^c$ belonging to the class $c$ according to the weights $w_j^c$ of the feature maps contributing to class $c$. We simplify the $\bm{A}^c$ as $\bm{A}$ in our binary classification task, denoted as,
\begin{equation}
    \bm{A}= ReLU(\sum_{j} w_j \bm{F}_j^{(4)}) 
\end{equation}
where $w_j$ is the element of weight matrix in the MLP layers, and $ReLU$ function removes the negative activation. Finally, we apply the min-max normalization to scale the values of $\bm{A}$ within $(0,1)$.

\subsection{CAM Adjustment}
 However, the derived CAM above is usually focused on part regions of the object, as the single label classification task does not enforce the model to attend to all parts of the object.
 %
 Thus, our goal is to help the network to capture the regions with the same appearance or textures to tightly cover this object.

 \noindent\textbf{Feature Contrast.}
As CAM is the sum of weighted feature maps, it is intuitive to improve the quality of the feature maps to get a precision CAM. We observe that the derived CAM usually focuses on discriminative regions, which cannot cover the whole region of dropping liquid. Thus, we primarily use the contrastive strategy to supervise  feature maps to force the network to gather foreground and background regions, respectively.

Give the activation map $\bm{A}$, we index the location $i$ which makes $\bm{A}\geq\epsilon$ and retrieve the equal-sized feature maps $\bm{F}^{(4)} $ to get the foreground local features $\bm{f}_i^{fg} \in \mathbb{R}^{1 \times d}$ as in Fig.~\ref{fig:pipeline}; $d$ is the dimension of the feature maps $\bm{F}^{(4)}$. By analogy, the background local features are denoted as $\bm{f}_j^{bg} \in \mathbb{R}^{1 \times d}$, where $\bm{A}\textless \epsilon$. Then we aim to narrow the distance between foreground local features, and push apart the distance between foreground-background ones. We use the cosine similarity metric $sim(\cdot, \cdot)$ to measure the distance between each location of feature maps.
For the positive contrastive loss, we aim to pull the features from similar regions together, supervised as, 
\begin{equation}
    \mathcal{L}_{pos} = -\frac{1}{m^2}\sum_{i=1}^m\sum_{i=1}^m log(sim(\bm{f}_i^{fg}, \bm{f}_i^{fg}))
\end{equation}

We also encourage the network to push apart the foreground and background features to obtain the more complete regions belonging to the actual flowing liquid.
\begin{equation}
    \mathcal{L}_{neg} = -\frac{1}{mn}\sum_{i=1}^m\sum_{j=1}^n log(1-sim(\bm{f}_i^{fg}, \bm{f}_j^{bg}))
\end{equation}


where $m$ is the number of indexed local features $\bm{f}_i^{fg}$, and $n$ is that of $\bm{f}_j^{bg}$. Finally, the overall loss is the sum of $\mathcal{L}_{cls}$, $\mathcal{L}_{neg}$ and $\mathcal{L}_{pos}$, which is formulated as
\begin{equation}
    \mathcal{L} = \mathcal{L}_{cls} + \mathcal{L}_{pos} + \mathcal{L}_{neg}
\end{equation}

\subsection{3D Modeling of Flowing Liquid}

\begin{figure}
    \centering
    \includegraphics[width=1\linewidth]{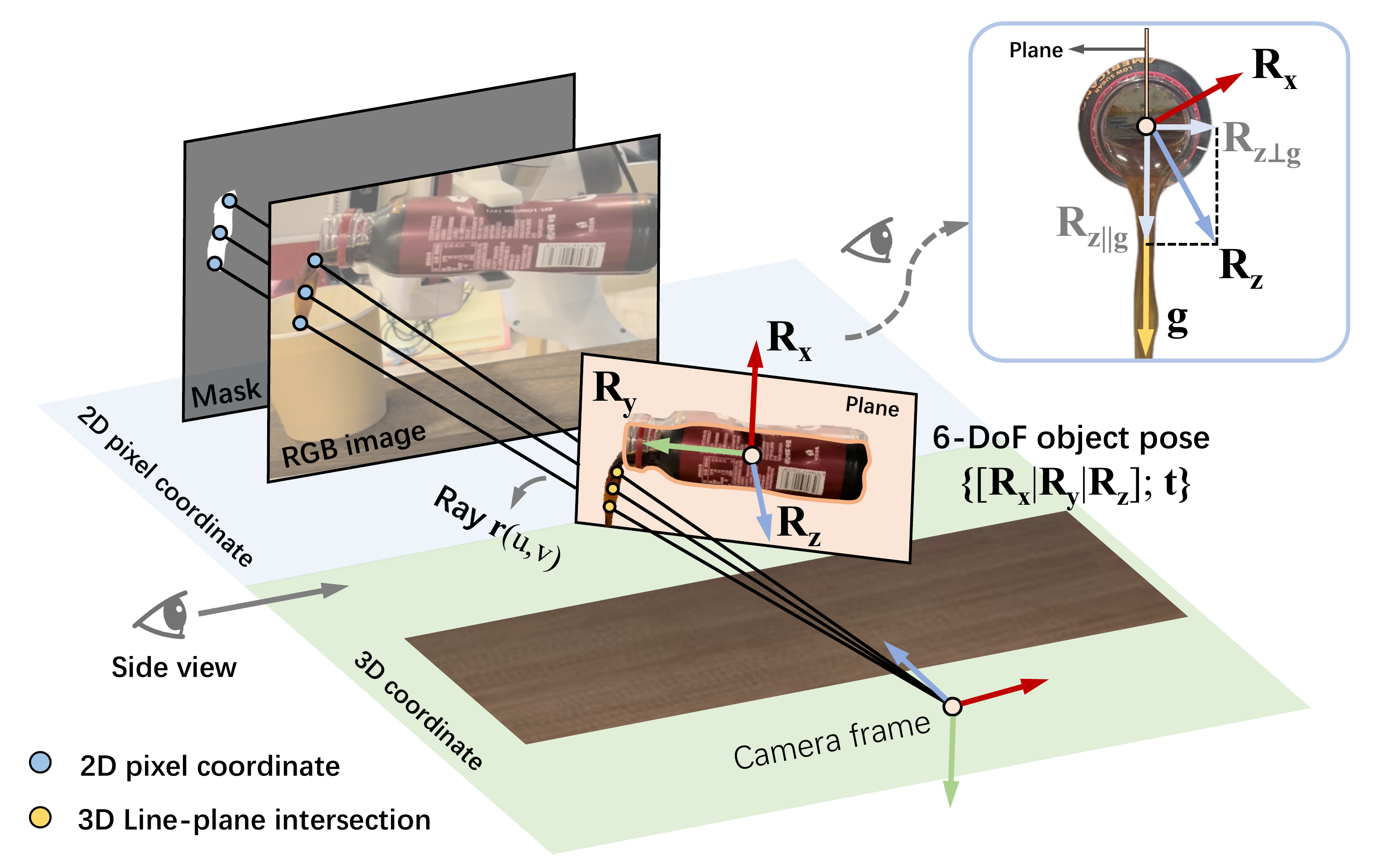}
    \caption{3D modeling of the liquid. For better illustration, we exchange the position of 3D coordinate and 2d pixel coordinate. }
    \label{fig:modeling}
    \vspace{-0.15in}
\end{figure}

 We calculate the region mask $\mathcal{M}$ of liquid from the CAM where $\bm{A}\geq \sigma$. Due to the transparency of the liquid, the depth camera cannot well calculate the depth of liquid pixels. Thus, we further take the container pose to model the approximate depth of the liquid as in Fig.~\ref{fig:modeling}.
The pose $\{\bm{R}=[\bm{R}_x|\bm{R}_y|\bm{R}_z]; \bm{t}\}$ of source container in camera frame is acquired by category-level pose estimator, \ie, SAR-Net~\cite{lin2022sar}. 
Then, the plane aligned with gravity and orientation of container’s bottleneck (the green vector $\bm{R}_y$) can be formulated as, 
\begin{equation}
\begin{cases}
    (\bm{p}-\bm{t})^\top \cdot \bm{R}_{z\perp g} = 0\\
    \bm{R}_z = \bm{R}_{z\perp g} + \bm{R}_{z||g} \\
\end{cases}
\end{equation}

where $\bm{p}=[x,y,z]^\top$ is the point on the plane, and $\bm{t}$ is the estimated 3D center of the source container in the camera frame. $\bm{R}_{z\perp g}$ and $\bm{R}_{z||g}$ are the components of $\bm{R}_{z}$ that are vertical and parallel to gravity vector $\bm{g}$ in the camera frame, respectively.
Given the camera intrinsic $\bm{K}$, the discrete ray $\bm{r}(u,v)$ emitted from the camera is formulated as,
\begin{equation}
\begin{cases}
    \bm{r}(u,v) =\bm{o} + \kappa\bm{d}, \kappa \in \mathbb{R}\\
    \bm{d} = \bm{K}^{-1} \cdot [u,v,1]^\top
\end{cases}
\end{equation}

where $(u, v)$ is the pixel coordinate. $\kappa$ is a scalar, which makes it possible to define any position along the ray. $\bm{o}$ and $\bm{d}$ indicate the origin and direction of the ray, respectively. 
Finally, we calculate the 3D intersection position $\bm{p}$ of the ray and the plane.
\begin{equation}
    \bm{p} = \kappa\bm{d}, \kappa = \frac{\bm{t}^\top \cdot \bm{R}_{z\perp g}}{\bm{d}^\top \cdot \bm{R}_{z\perp g}}
\end{equation}

With this processing, we back-project  2D pixel $(u, v)$  of liquid region from pixel coordinate into the actual 3D coordinate space. Finally, given the pixel $(u, v) $ inside the mask $\mathcal{M}$, we can get the approximate 3D point cloud of the liquid $\bm{P}=\{\bm{p}(u,v)|(u,v)\in \mathcal{M}\}$.
Actually, we post-process the mask $\mathcal{M}$ by using the skeletonization algorithm~\cite{zhang1984fast}.
Although the 3D liquid modeling is based on  gravitational assumption, further real-world robotics experiment in Tab.~\ref{tab:comparison_robotics} shows that such an assumption is reasonable and robust enough to serve the robot as visual feedback for control.

\section{Experiment}
\noindent\textbf{Dataset.}
\noindent\textit{(1) PourIt! Dataset.} 
This dataset has 4064 RGB images recorded by a Kinect Azure camera with different backgrounds (Fig.~\ref{fig:dataset_example}). We split them into 3354 for training and 710 for testing. The testing set has 374 images of familiar scenes and 336 images of new scenes. We labeled only the testing set for evaluation since annotating every liquid pixel is tough. It includes images of a person pouring liquid, a robot pouring liquid using pose estimation, and water flowing from various taps in a bathroom. It is a test bed for self-supervised methods.
\textit{(2) Liquid Dataset.} It contains 648 pouring trials on the dual-arm Baxter robot. The robot rotates its wrist to pour the liquid from the source container to the target one. 
Each trial is generated by controlling the six variables: (1) left or right arm that grasps the source container; (2) different source containers; (3) different target containers; (4) the initial volume of the liquid in source container; (5) pouring trajectory; and (6) the proportion of the person appearing in the background.
We split the sequences of each scene into images with liquid or non-liquid. The images are cropped at the center into the shape of $300 \times 300$ to filter out the irrelevant background. The details of the data split are illustrated in our Appendix. We finally sample 6387 training images and 1598 testing images.


\noindent\textbf{Evaluation Metric.}
To evaluate the segmentation accuracy of the segmented mask generated by different methods, we calculate the Intersection over Union (IoU) between the ground-truth segmentation labels and predicted ones and the mean IoU(\%). For our method and EPS~\cite{lee2021railroad}, we generate the segmented mask using threshold $\sigma$ of 0.5 and 0.7, where $\bm{A} \geq \sigma$. And the mIoU = $\sum_{\sigma=0.5,0.7}$ mIoU$(\sigma)/2$.

\noindent\textbf{Implementation Details.} We use six 2080Ti GPUs for training with a batch size of 2 on each GPU. It takes $14000$ iterations (nearly $1.5h$) to converge on the PourIt! dataset while $18000$ iterations on the Liquid dataset (nearly $2h$). We use the AdamW optimizer with the initial learning rate of $6\times 10^{-5}$ for the backbone parameters. The learning rate of other parameters is ten times the parameters of backbone, \ie, $6\times 10^{-4}$. We also adopt the data augmentation with random re-scaling with a range of [1.0, 1.1], random horizontally flipping, and random cropping with a cropping size of $512\times512$, respectively. Our model is warmed up by only using $\mathcal{L}_{cls}$ for the early 2000 iterations to get an initial CAM. We set $\epsilon=0.7$ in our experiment. 

\subsection{Comparison to Baseline}
\begin{figure}
    \centering
    \includegraphics[width=1\linewidth]{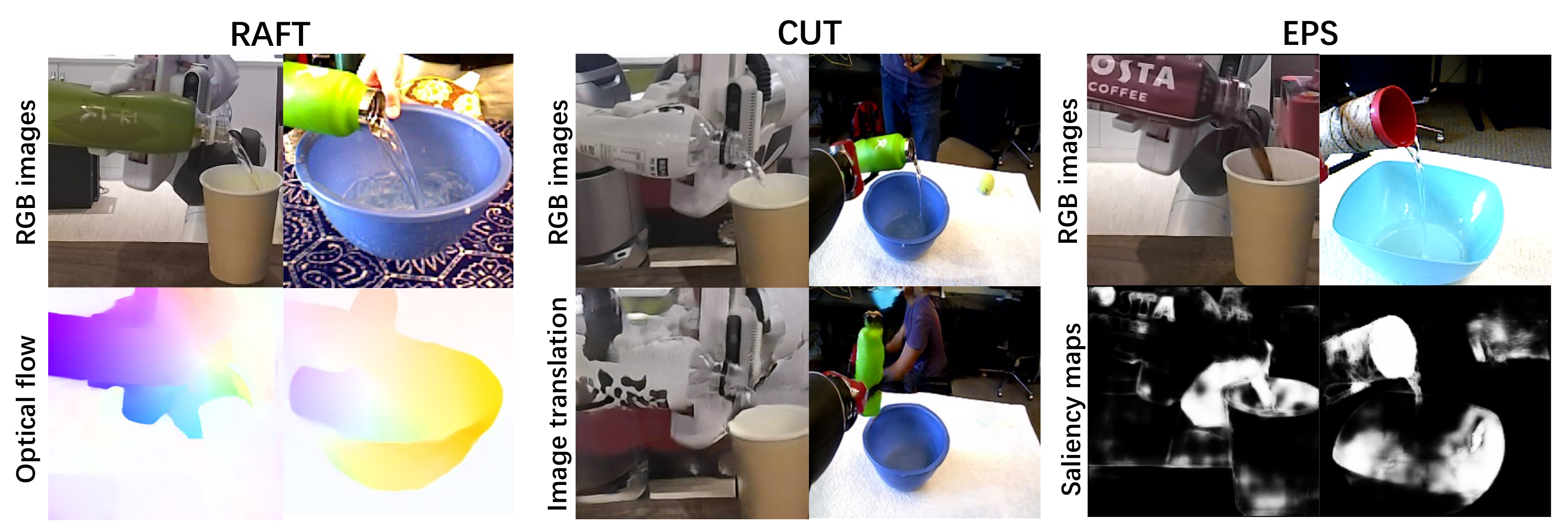}
    \caption{Qualitative intermediate results of CUT, RAFT and EPS methods. For RAFT method, per-pixel displacement vector is colorized in different colors (left). For CUT, we show the image-to-image translation results of transferring the liquid images into non-liquid ones (center). For EPS, we visualize the saliency maps used for supervision (right).}
    \label{fig:raft_cut_eps}
\end{figure}

\noindent\textbf{Baseline.} Liquid~\cite{schenck2016towards} is the first supervised method for liquid segmentation, we implement this method as the upper bound to help compare performance. 
Then, we compare another three types of baseline methods, including (1) optical flow method (RAFT~\cite{teed2020raft}), (2) image-to-image translation method (CUT~\cite{park2020contrastive}) and (3) CAM-based methods (EPS~\cite{lee2021railroad} and AFA~\cite{ru2022learning}).
Particularly, RAFT is used to detect pixel displacement between frames from the sequences of images for liquid region detection as in Fig.~\ref{fig:raft_cut_eps} (left column). 
 Similar to ~\cite{narasimhan2022self}, CUT is utilized to transfer the liquid images into non-liquid images as in Fig.~\ref{fig:raft_cut_eps} (right column), and then use background subtraction algorithm to estimate the visual difference between the original and non-liquid translation images. AFA and our method all use the MiT-B1 backbones.

\noindent\textbf{Results on PourIt Dataset.}
We compare several baseline methods in our proposed PourIt! dataset and report the results in terms of mIoU in Tab.~\ref{tab:comparison_pourit}. We observe that our method has a significantly outperforms the state-of-the-art WSSS method AFA, by a large margin of 10.1\% on seen images and 10.8\% on unseen images. This also validates the generalize ability of our proposed method. Although AFA uses the pixel-adaption refinement strategy, but the liquid is very different with the solid object which usually have salient boundaries. Such a strategy fails to get a better pseudo labels for segmentation training.

The RAFT, CUT and EPS methods cannot well handle the images with liquid on PourIt! dataset, thus we also provide the qualitative results to illustrate the failed cases as in Fig.~\ref{fig:raft_cut_eps}. Concretely, RAFT~\cite{teed2020raft} fails to segment the liquid due to the background movement. Thus, the optical flow method is only suitable for simple static scenes where the containers and background are relatively stationary. Nevertheless, the translation images from CUT method~\cite{park2020contrastive} have poor results due to the various contextual information across images. It simultaneously erases the foreground but also unexpectedly modifies the background context, making calculating the mask of the liquid region unavailable. EPS~\cite{lee2021railroad} has poor performance as the saliency maps cannot focus on the actual liquid region for supervision.


\noindent\textbf{Results on Liquid Dataset.}
The comparison results on the Liquid dataset are summarized in Tab.~\ref{tab:comparison_liquid} and Fig.~\ref{fig:raft_cut_eps}. 
Our method achieves 54.0\% mIoU on the test set, and surpasses AFA method by 4.6\%. The RAFT and CUT methods are still failed in this dataset, by analogy to the aforementioned analyse of the failed cases in PourIt! dataset. We also adopt the fully-supervised method Liquid~\cite{schenck2016towards} as the oracle approach, thus to help us understand the upper bound of the performance of liquid segmentation.

\noindent \textit{Discussion about limitations of Liquid dataset}. The annotations in Liquid dataset contain liquid out-flowed from bottleneck, and liquid stayed in the container.
While the original purpose of the dataset~\cite{schenck2016towards} was to estimate the volume in the container, we thus provide experimental results that serve as a test bed to validate our proposed method.
We acknowledge that the model trained on this dataset cannot directly apply to control the real-world robot for accurate pouring, but our approach presents a valuable contribution to the field. Please refer to  Appendix for  more details.





\begin{figure*}[h]
    \centering
    \includegraphics[width=1\linewidth]{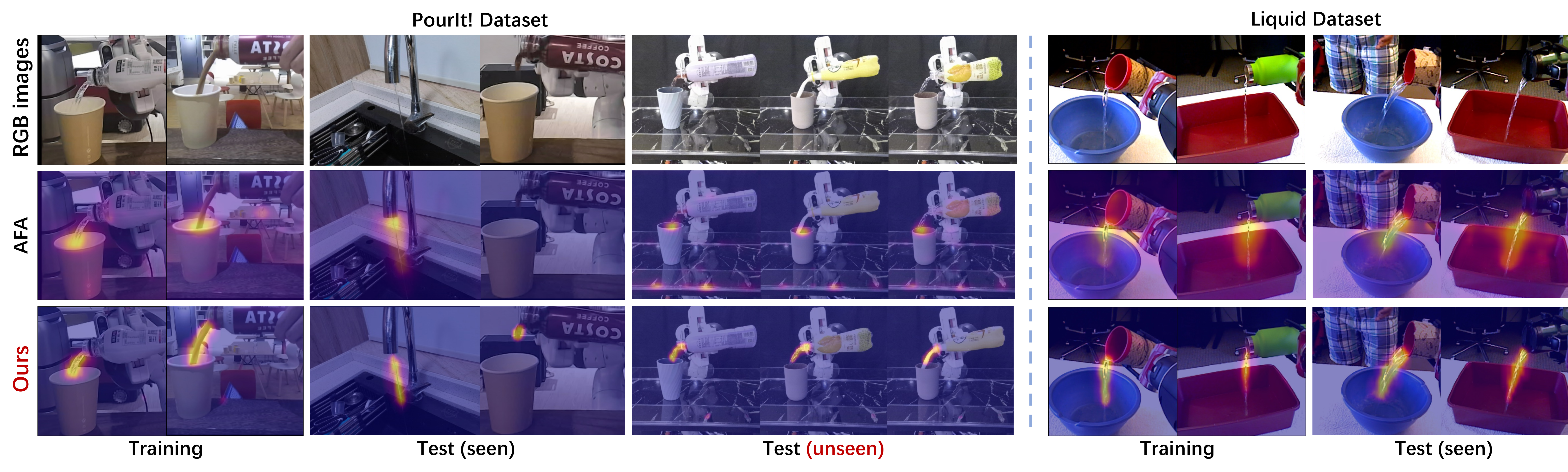}
    \vspace{-0.2in}
    \caption{CAM visualization  of different methods. Note that the quality of AFA's CAM is similar to its segmented mask results, thus we only visualize  CAM but not AFA segmented masks~\cite{ru2022learning} for consistency of visual contrast. The comparison results show that our CAM results more tightly focus on complete liquid region, with well alignment of low-level boundaries, and generalize across novel scenes. 
    \label{fig:vis_dataset}}
    \vspace{-0.15in}
\end{figure*}

\noindent\textbf{Qualitative Results.}
We also compare our CAM results with the baseline on two datasets as shown in Fig.~\ref{fig:vis_dataset}. Obviously, our method outputs more accurate CAM covering the liquid regions, especially for the thin stream of liquid. The results on novel scenes show that our method is robustly generalized to different kinds of liquid, containers, and background contexts.

\begin{table}[]
\small
\centering
\caption{Segmentation results on PourIt! dataset. We report the performance between our method and the baseline methods in metric of mIoU(\%). \textit{Sup.} denotes supervision type. $\mathcal{I}$: image-level labels. $\mathcal{S}$: saliency maps. $\dag$ denotes our implementation. \label{tab:comparison_pourit}} 
\begin{tabular}{l|c|c|cc}
\toprule[1pt]
\textit{Method} & \textit{Sup.} & \textit{Backbone} & \textit{seen} & \textit{unseen} \\ \midrule[0.5pt]
RAFT~\cite{teed2020raft}~\textsubscript{ECCV'20} &  --  &  ResNet  &  9.4  & 10.8 \\

CUT$^\dag$ ~\cite{park2020contrastive}~\textsubscript{ECCV'20}   & $\mathcal{I}$    &  ResNet &  38.1 &  24.3 \\

EPS$^\dag$~\cite{lee2021railroad}~\textsubscript{CVPR'21} &  $\mathcal{I}+\mathcal{S}$  &  ResNet38  & 47.2  & 42.0 \\
AFA$^\dag$~\cite{ru2022learning}~\textsubscript{CVPR'22}        & $\mathcal{I}$    & MiT-B1   & 55.8   & 52.8 \\
Ours         & $\mathcal{I}$    & MiT-B1   &    \textbf{65.9}   & \textbf{63.6} \\ \bottomrule[1pt]
\end{tabular}
\vspace{-0.15in}
\end{table}

\begin{table}[]
\small
\centering
\caption{Segmentation results on Liquid dataset. We report  performance between ours and baseline methods in metric of mIoU(\%). \textit{Sup.} denotes supervision type. $\mathcal{F}$: full supervision; $\mathcal{I}$: image-level labels. $\mathcal{S}$: saliency maps. $\dag$ denotes our implementation. \label{tab:comparison_liquid}}
\begin{tabular}{l|c|c|c}
\toprule[1pt]
\textit{Method} & \textit{Sup.} & \textit{Backbone} & \textit{seen} \\ \midrule[0.5pt]
Liquid$^\dag$~\cite{schenck2016towards} ~\textsubscript{Upper bound}  & $\mathcal{F}$    & FCN   &  69.0    \\ \midrule[0.5pt]
RAFT~\cite{teed2020raft}~\textsubscript{ECCV'20} &   --   &   ResNet       &   5.5      \\ 
CUT$^\dag$~\cite{park2020contrastive}~\textsubscript{ECCV'20}          & $\mathcal{I}$    & ResNet  &   38.2   \\
EPS$^\dag$~\cite{lee2021railroad}~\textsubscript{CVPR'21} &  $\mathcal{I}+\mathcal{S}$  &  ResNet38  &  41.6  \\
AFA$^\dag$~\cite{ru2022learning}~\textsubscript{CVPR'22}   & $\mathcal{I}$    & MiT-B1   &    49.4    \\
Ours         & $\mathcal{I}$    & MiT-B1   &  \textbf{54.0} \\ \bottomrule[1pt]
\end{tabular}
\vspace{-0.15in}
\end{table}

\subsection{Ablation Study and Analysis}
We perform ablation studies to validate the contribution of each loss employed in our method. We compare the full model and its variants which disable the loss ($\mathcal{L}_{pos}$, $\mathcal{L}_{neg}$ and $\mathcal{L}_{pos+neg}$).
We test these variants on the PourIt! dataset. We summarize the results in Tab.~\ref{tab:comparison_ablation_loss}. 

It is evident that $\mathcal{L}_{neg}$ is the most significant among them, as it helps the network better distinguish the foreground and background regions. It also helps the network generalize to different liquids, containers and backgrounds, thus maintaining the performance, \ie, 65.9\% on seen images vs. 63.9\% on unseen images. Interestingly, single use of $\mathcal{L}_{pos}$ negatively influences the performance of the network. One explanation is that the highly responsive area of the imperfect initial CAM may contains foreground and background pixels; and only using $\mathcal{L}_{pos}$ will guide the network to be improperly optimized to pull together the actual foreground and background features. This will result in the chaos of distinguishing the foreground and background pixels, especially on the unseen images, \ie, 46.7\% on seen images vs. 36.7\% on unseen images. 
Especially, the combination of $\mathcal{L}_{pos}$ and $\mathcal{L}_{neg}$ encourage the network to differentiate clear partition between the foreground and background pixels, even though the initial regions which CAM puts focus on may contain some pixels from background. The final results reveal that our well-designed losses benefit network learning and help boost the final performance. 

\subsection{Generalization Analysis} 
For training images in PourIt! dataset, we use three types of source and target containers, respectively. Moreover, the liquid contains \textit{transparent water}, \textit{coffee}, and \textit{tea}. We use another three types of source and target containers for unseen test images. Furthermore, the liquid contains \textit{milk}, \textit{juice of grape}, and \textit{melon}. 
Surprisingly, our trained model could be naturally generalized well across generalized to images of different liquids, containers, and even the background as in Fig.~\ref{fig:vis_dataset} (column 3). This verifies that our model is well-trained to learn the category-level feature representations of liquid with different appearances.
Thus, this category-level generalization ability further endows the feasibility of our model in different real-world scenes.

\begin{table}[]
\small
\centering
\setlength\tabcolsep{4pt}
\caption{Ablation study for different losses on PourIt! dataset under the metric of mIoU(\%).
\label{tab:comparison_ablation_loss}}
\begin{tabular}{c|cccc}
\toprule[1pt]
Variants & w/o $\mathcal{L}_{neg+pos}$ & w/o $\mathcal{L}_{neg}$ &  w/o $\mathcal{L}_{pos}$  & Full \\ \midrule[0.5pt]
\textit{seen} &  61.6  &  46.7  & 60.6  &  \textbf{65.9}   \\
\textit{unseen}  &  59.2  & 36.7 & 63.2 & \textbf{63.6}  \\ \bottomrule[1pt]
\end{tabular}
\vspace{-0.15in}
\end{table}

\subsection{Robotics Experiments}
\noindent\textbf{Hardware Experimental Setup.}
We use a Franka-Emika Panda 7-DoF robotic arm~\cite{franka} with a parallel jaw gripper to perform object grasping and manipulation for pouring tasks. The Azure Kinect camera~\cite{kinect} provides RGB-D sensing, which is mounted on the tripod opposite the robot's workspace. This camera is also calibrated to the robot’s base frame. In our experiment, three desktop computers were utilized: one computer for the real-time control of Franka, two for real-time visual perception, \ie, the pose estimation network is distributed on a laptop with an NVIDIA RTX 3070 GPU which runs at 20Hz per object, and the liquid perception network is deployed on a desktop with an NVIDIA RTX 2070 GPU which runs at 15Hz per frame. 

\noindent\textbf{Task Description and Evaluation Metric.}
We design two kinds of tasks for robotic pouring, including static scenes of pouring liquid into a stationary position target container and dynamic scenes of pouring liquid into a moving target container. For the moving target container, the experimenter will hold the container in hand with linear or random motion. We use six groups of different source and target containers for testing. 
Successful pouring means the liquid is accurately poured into the target container without spilling.
We try 15 attempts for each type of scene and record the total number of successes. More details in Appendix please.

\begin{figure}[ht]
    \centering
    \includegraphics[width=1\linewidth]{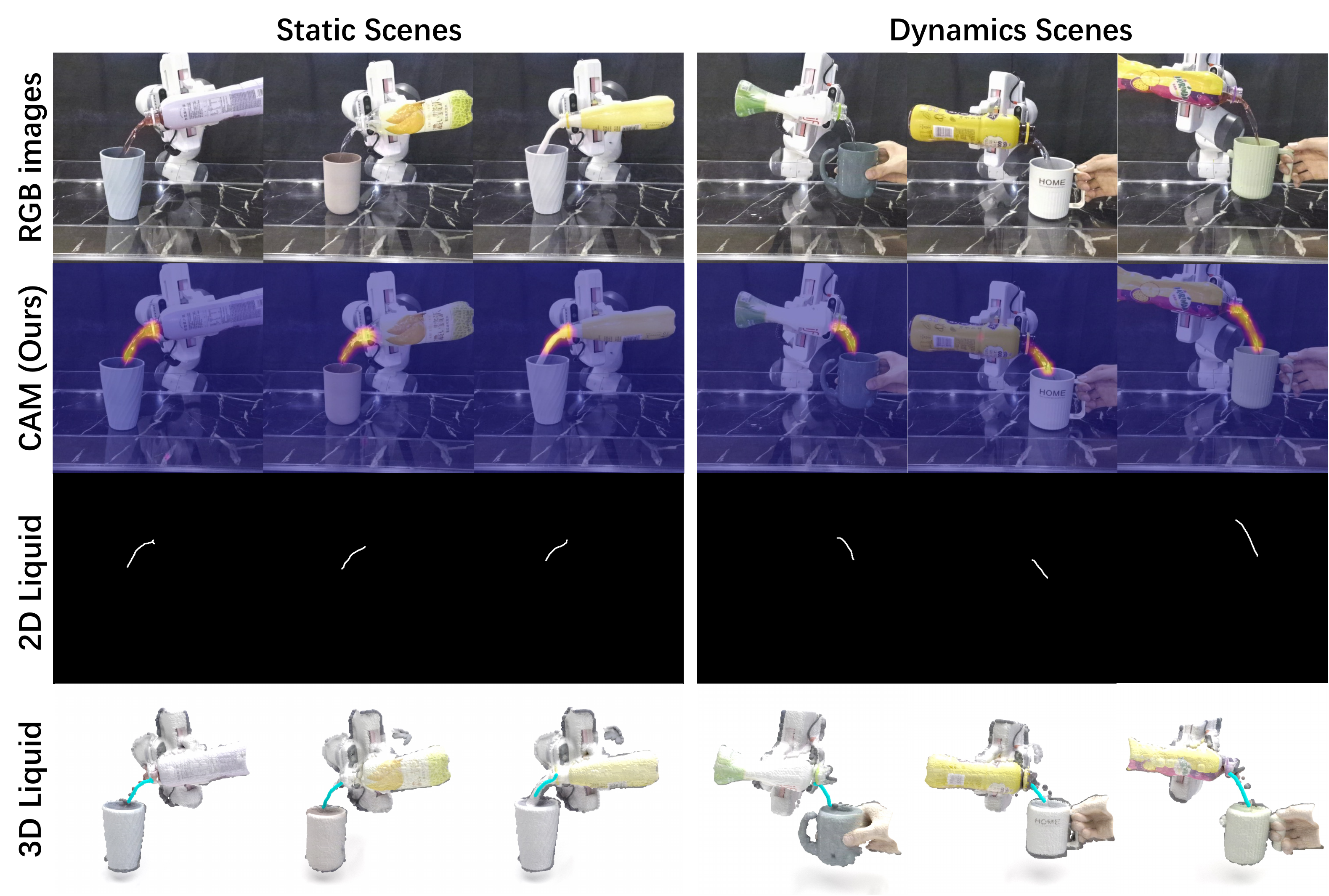}
    \vspace{-0.15in}
    \caption{Visualization of the output results of our PourIt! framework. We show the original RGB images, predicted CAM, skeletonized 2D liquid trajectory and estimated 3D point cloud of liquid (cyan points). 
    \label{fig:vis_robotics}}
    \vspace{-0.15in}
\end{figure}

\noindent\textbf{Results and Analysis.}
We report the success rate of the pouring task in Tab.~\ref{tab:comparison_robotics}, which proves the efficacy of our method. The method of Lin~\etal~\cite{lin2022sar} only uses the estimated object pose and size for calculating container-to-container distance as the initial pouring point. Such a strategy sometimes fails due to an imperfect initial pouring point, as the transparent bottleneck results in inaccurate object size estimation. Our method only use object pose information to calculate the liquid-to-container distance as feedback to adjust the pouring point, which is less influenced by the inaccurate size estimation. However, AFA method performs poorly than \cite{lin2022sar}, because the inaccurately estimated mask of AFA provides the wrong visual feedback to adjust the pouring point, making the liquid spill out of the target container instead. 
Figure~\ref{fig:vis_robotics} demonstrates the visualization of intermediate results from the experiments. Significantly, we recover the 3D point cloud of liquid due to the unavailable depth measurement for transparent liquid. We observe that the depth of  milk liquid is still roughly measurable, and our reconstructed point cloud (cyan points in column 3) is aligned well with the shape recovered from the sensor-measured depth. This verifies that our method can accurately reconstruct the shape of a stream of liquid. 

\begin{table}[] 
\small
\centering
\caption{Success rate of robotics pouring. `Static' indicates that the target container is placed statically on the desktop. `Dym.(L)' means the target container is held on experimenter's hand moving by linear motion, and `Dym.(R)' denotes the random motion. \label{tab:comparison_robotics}}
\begin{tabular}{c|ccc}
\toprule[1pt]
\multirow{2}{*}{$Scenes$} & \multicolumn{3}{c}{$Success\quad Rate (\%)$} \\
                        & $Static$  & $Dym.(L)$  & $Dym.(R)$  \\ \midrule[0.5pt]
Lin~\etal~\cite{lin2022sar}     &   66.7    &    53.3    &  40.0            \\
AFA~\cite{ru2022learning}         &   60.0    &    46.7   &  33.3                \\
Ours                    &  93.3  &  73.3   &  60.0            \\ \bottomrule[1pt]
\end{tabular}
\end{table}



\section{Conclusion}
This paper presents a simple yet effective framework PourIt! for  visual closed-loop robotic pouring task. It only takes image-level labels and uses the CAM technique for liquid segmentation. We also design a feature contrast strategy to improve the quality of the CAM, thus better localizing the potential liquid regions. Then, the container's pose is further utilized to facilitates the 3D point cloud recovery. Finally, we take the liquid-to-container distance as the feedback to control the robot. In future work, we are interested in exploring the recovered 3D liquid shape for more different pouring-related tasks. 

{\small
\bibliographystyle{ieee_fullname}
\bibliography{egbib}

\begin{thebibliography}{10}\itemsep=-1pt

\bibitem{kinect}
{Azure Kinect DK}.
\newblock \url{https://azure.microsoft.com/en-us/products/kinect-dk}.

\bibitem{franka}
{Franka Emika Panda}.
\newblock \url{https://www.franka.de}.

\bibitem{ahn2018learning}
Jiwoon Ahn and Suha Kwak.
\newblock Learning pixel-level semantic affinity with image-level supervision
  for weakly supervised semantic segmentation.
\newblock In {\em Proceedings of the IEEE conference on computer vision and
  pattern recognition}, pages 4981--4990, 2018.

\bibitem{bearman2016s}
Amy Bearman, Olga Russakovsky, Vittorio Ferrari, and Li Fei-Fei.
\newblock What’s the point: Semantic segmentation with point supervision.
\newblock In {\em Computer Vision--ECCV 2016: 14th European Conference,
  Amsterdam, The Netherlands, October 11--14, 2016, Proceedings, Part VII 14},
  pages 549--565. Springer, 2016.

\bibitem{chen2022class}
Zhaozheng Chen, Tan Wang, Xiongwei Wu, Xian-Sheng Hua, Hanwang Zhang, and
  Qianru Sun.
\newblock Class re-activation maps for weakly-supervised semantic segmentation.
\newblock In {\em Proceedings of the IEEE/CVF Conference on Computer Vision and
  Pattern Recognition}, pages 969--978, 2022.

\bibitem{cheng2021per}
Bowen Cheng, Alex Schwing, and Alexander Kirillov.
\newblock Per-pixel classification is not all you need for semantic
  segmentation.
\newblock {\em Advances in Neural Information Processing Systems},
  34:17864--17875, 2021.

\bibitem{chitta2012moveit}
Sachin Chitta, Ioan Sucan, and Steve Cousins.
\newblock Moveit![ros topics].
\newblock {\em IEEE Robotics \& Automation Magazine}, 19(1):18--19, 2012.

\bibitem{dai2015boxsup}
Jifeng Dai, Kaiming He, and Jian Sun.
\newblock Boxsup: Exploiting bounding boxes to supervise convolutional networks
  for semantic segmentation.
\newblock In {\em Proceedings of the IEEE international conference on computer
  vision}, pages 1635--1643, 2015.

\bibitem{ishikawa2021audio}
Reina Ishikawa, Yuichi Nagao, Ryo Hachiuma, and Hideo Saito.
\newblock Audio-visual hybrid approach for filling mass estimation.
\newblock In {\em International Conference on Pattern Recognition}, pages
  437--450. Springer, 2021.

\bibitem{jiang2019integral}
Peng-Tao Jiang, Qibin Hou, Yang Cao, Ming-Ming Cheng, Yunchao Wei, and Hong-Kai
  Xiong.
\newblock Integral object mining via online attention accumulation.
\newblock In {\em Proceedings of the IEEE/CVF international conference on
  computer vision}, pages 2070--2079, 2019.

\bibitem{khoreva2017simple}
Anna Khoreva, Rodrigo Benenson, Jan Hosang, Matthias Hein, and Bernt Schiele.
\newblock Simple does it: Weakly supervised instance and semantic segmentation.
\newblock In {\em Proceedings of the IEEE conference on computer vision and
  pattern recognition}, pages 876--885, 2017.

\bibitem{kim2021discriminative}
Beomyoung Kim, Sangeun Han, and Junmo Kim.
\newblock Discriminative region suppression for weakly-supervised semantic
  segmentation.
\newblock In {\em Proceedings of the AAAI Conference on Artificial
  Intelligence}, volume~35, pages 1754--1761, 2021.

\bibitem{kolesnikov2016seed}
Alexander Kolesnikov and Christoph~H Lampert.
\newblock Seed, expand and constrain: Three principles for weakly-supervised
  image segmentation.
\newblock In {\em Computer Vision--ECCV 2016: 14th European Conference,
  Amsterdam, The Netherlands, October 11--14, 2016, Proceedings, Part IV 14},
  pages 695--711. Springer, 2016.

\bibitem{lee2021railroad}
Seungho Lee, Minhyun Lee, Jongwuk Lee, and Hyunjung Shim.
\newblock Railroad is not a train: Saliency as pseudo-pixel supervision for
  weakly supervised semantic segmentation.
\newblock In {\em Proceedings of the IEEE/CVF conference on computer vision and
  pattern recognition}, pages 5495--5505, 2021.

\bibitem{liang2019making}
Hongzhuo Liang, Shuang Li, Xiaojian Ma, Norman Hendrich, Timo Gerkmann, Fuchun
  Sun, and Jianwei Zhang.
\newblock Making sense of audio vibration for liquid height estimation in
  robotic pouring.
\newblock In {\em 2019 IEEE/RSJ International Conference on Intelligent Robots
  and Systems (IROS)}, pages 5333--5339. IEEE, 2019.

\bibitem{lin2016scribblesup}
Di Lin, Jifeng Dai, Jiaya Jia, Kaiming He, and Jian Sun.
\newblock Scribblesup: Scribble-supervised convolutional networks for semantic
  segmentation.
\newblock In {\em Proceedings of the IEEE conference on computer vision and
  pattern recognition}, pages 3159--3167, 2016.

\bibitem{lin2022sar}
Haitao Lin, Zichang Liu, Chilam Cheang, Yanwei Fu, Guodong Guo, and Xiangyang
  Xue.
\newblock Sar-net: Shape alignment and recovery network for category-level 6d
  object pose and size estimation.
\newblock In {\em Proceedings of the IEEE/CVF Conference on Computer Vision and
  Pattern Recognition}, pages 6707--6717, 2022.

\bibitem{liu2021va2mass}
Qi Liu, Fan Feng, Chuanlin Lan, and Rosa~HM Chan.
\newblock Va2mass: Towards the fluid filling mass estimation via integration of
  vision and audio learning.
\newblock In {\em International Conference on Pattern Recognition}, pages
  451--463. Springer, 2021.

\bibitem{miller2004graspit}
Andrew~T Miller and Peter~K Allen.
\newblock Graspit! a versatile simulator for robotic grasping.
\newblock {\em IEEE Robotics \& Automation Magazine}, 11(4):110--122, 2004.

\bibitem{narasimhan2022self}
Gautham Narasimhan, Kai Zhang, Ben Eisner, Xingyu Lin, and David Held.
\newblock Self-supervised transparent liquid segmentation for robotic pouring.
\newblock In {\em 2022 International Conference on Robotics and Automation
  (ICRA)}, pages 4555--4561. IEEE, 2022.

\bibitem{papandreou2015weakly}
George Papandreou, Liang-Chieh Chen, Kevin~P Murphy, and Alan~L Yuille.
\newblock Weakly-and semi-supervised learning of a deep convolutional network
  for semantic image segmentation.
\newblock In {\em Proceedings of the IEEE international conference on computer
  vision}, pages 1742--1750, 2015.

\bibitem{park2020contrastive}
Taesung Park, Alexei~A Efros, Richard Zhang, and Jun-Yan Zhu.
\newblock Contrastive learning for unpaired image-to-image translation.
\newblock In {\em Computer Vision--ECCV 2020: 16th European Conference,
  Glasgow, UK, August 23--28, 2020, Proceedings, Part IX 16}, pages 319--345.
  Springer, 2020.

\bibitem{ru2022learning}
Lixiang Ru, Yibing Zhan, Baosheng Yu, and Bo Du.
\newblock Learning affinity from attention: end-to-end weakly-supervised
  semantic segmentation with transformers.
\newblock In {\em Proceedings of the IEEE/CVF Conference on Computer Vision and
  Pattern Recognition}, pages 16846--16855, 2022.

\bibitem{schenck2016towards}
Connor Schenck and Dieter Fox.
\newblock Towards learning to perceive and reason about liquids.
\newblock In {\em International Symposium on Experimental Robotics}, pages
  488--501. Springer, 2016.

\bibitem{schenck2017reasoning}
Connor Schenck and Dieter Fox.
\newblock Reasoning about liquids via closed-loop simulation.
\newblock {\em arXiv preprint arXiv:1703.01656}, 2017.

\bibitem{schenck2017visual}
Connor Schenck and Dieter Fox.
\newblock Visual closed-loop control for pouring liquids.
\newblock In {\em 2017 IEEE International Conference on Robotics and Automation
  (ICRA)}, pages 2629--2636. IEEE, 2017.

\bibitem{schenck2018perceiving}
Connor Schenck and Dieter Fox.
\newblock Perceiving and reasoning about liquids using fully convolutional
  networks.
\newblock {\em The International Journal of Robotics Research},
  37(4-5):452--471, 2018.

\bibitem{selvaraju2017grad}
Ramprasaath~R Selvaraju, Michael Cogswell, Abhishek Das, Ramakrishna Vedantam,
  Devi Parikh, and Dhruv Batra.
\newblock Grad-cam: Visual explanations from deep networks via gradient-based
  localization.
\newblock In {\em Proceedings of the IEEE international conference on computer
  vision}, pages 618--626, 2017.

\bibitem{sun2019saliency}
Fengdong Sun and Wenhui Li.
\newblock Saliency guided deep network for weakly-supervised image
  segmentation.
\newblock {\em Pattern Recognition Letters}, 120:62--68, 2019.

\bibitem{sun2021ecs}
Kunyang Sun, Haoqing Shi, Zhengming Zhang, and Yongming Huang.
\newblock Ecs-net: Improving weakly supervised semantic segmentation by using
  connections between class activation maps.
\newblock In {\em Proceedings of the IEEE/CVF International Conference on
  Computer Vision}, pages 7283--7292, 2021.

\bibitem{teed2020raft}
Zachary Teed and Jia Deng.
\newblock Raft: Recurrent all-pairs field transforms for optical flow.
\newblock In {\em Computer Vision--ECCV 2020: 16th European Conference,
  Glasgow, UK, August 23--28, 2020, Proceedings, Part II 16}, pages 402--419.
  Springer, 2020.

\bibitem{wang2022improving}
Hengyi Wang, Chaoran Zhu, Ziyin Ma, and Changjae Oh.
\newblock Improving generalization of deep networks for estimating physical
  properties of containers and fillings.
\newblock In {\em ICASSP 2022-2022 IEEE International Conference on Acoustics,
  Speech and Signal Processing (ICASSP)}, pages 9147--9151. IEEE, 2022.

\bibitem{wei2017object}
Yunchao Wei, Jiashi Feng, Xiaodan Liang, Ming-Ming Cheng, Yao Zhao, and
  Shuicheng Yan.
\newblock Object region mining with adversarial erasing: A simple
  classification to semantic segmentation approach.
\newblock In {\em Proceedings of the IEEE conference on computer vision and
  pattern recognition}, pages 1568--1576, 2017.

\bibitem{wilson2019analyzing}
Justin Wilson, Auston Sterling, and Ming~C Lin.
\newblock Analyzing liquid pouring sequences via audio-visual neural networks.
\newblock In {\em 2019 IEEE/RSJ International Conference on Intelligent Robots
  and Systems (IROS)}, pages 7702--7709. IEEE, 2019.

\bibitem{xie2021segformer}
Enze Xie, Wenhai Wang, Zhiding Yu, Anima Anandkumar, Jose~M Alvarez, and Ping
  Luo.
\newblock Segformer: Simple and efficient design for semantic segmentation with
  transformers.
\newblock {\em Advances in Neural Information Processing Systems},
  34:12077--12090, 2021.

\bibitem{xompero2022audio}
Alessio Xompero, Yik~Lung Pang, Timothy Patten, Ahalya Prabhakar, Berk Calli,
  and Andrea Cavallaro.
\newblock Audio-visual object classification for human-robot collaboration.
\newblock In {\em ICASSP 2022-2022 IEEE International Conference on Acoustics,
  Speech and Signal Processing (ICASSP)}, pages 9137--9141. IEEE, 2022.

\bibitem{yamaguchi2016stereo}
Akihiko Yamaguchi and Christopher~G Atkeson.
\newblock Stereo vision of liquid and particle flow for robot pouring.
\newblock In {\em 2016 IEEE-RAS 16th International Conference on Humanoid
  Robots (Humanoids)}, pages 1173--1180. IEEE, 2016.

\bibitem{yao2021non}
Yazhou Yao, Tao Chen, Guo-Sen Xie, Chuanyi Zhang, Fumin Shen, Qi Wu, Zhenmin
  Tang, and Jian Zhang.
\newblock Non-salient region object mining for weakly supervised semantic
  segmentation.
\newblock In {\em Proceedings of the IEEE/CVF Conference on Computer Vision and
  Pattern Recognition}, pages 2623--2632, 2021.

\bibitem{zhang2021complementary}
Fei Zhang, Chaochen Gu, Chenyue Zhang, and Yuchao Dai.
\newblock Complementary patch for weakly supervised semantic segmentation.
\newblock In {\em Proceedings of the IEEE/CVF International Conference on
  Computer Vision}, pages 7242--7251, 2021.

\bibitem{zhang1984fast}
Tongjie~Y Zhang and Ching~Y. Suen.
\newblock A fast parallel algorithm for thinning digital patterns.
\newblock {\em Communications of the ACM}, 27(3):236--239, 1984.

\bibitem{zheng2021rethinking}
Sixiao Zheng, Jiachen Lu, Hengshuang Zhao, Xiatian Zhu, Zekun Luo, Yabiao Wang,
  Yanwei Fu, Jianfeng Feng, Tao Xiang, Philip~HS Torr, et~al.
\newblock Rethinking semantic segmentation from a sequence-to-sequence
  perspective with transformers.
\newblock In {\em Proceedings of the IEEE/CVF conference on computer vision and
  pattern recognition}, pages 6881--6890, 2021.

\bibitem{zhou2016learning}
Bolei Zhou, Aditya Khosla, Agata Lapedriza, Aude Oliva, and Antonio Torralba.
\newblock Learning deep features for discriminative localization.
\newblock In {\em Proceedings of the IEEE conference on computer vision and
  pattern recognition}, pages 2921--2929, 2016.

\end{thebibliography}
}

\end{document}